\pdfoutput=1
\documentclass[11pt]{article}

\usepackage[final]{acl}

\usepackage{times}
\usepackage{latexsym}
\usepackage[many]{tcolorbox}

\usepackage[T1]{fontenc}
\usepackage[utf8]{inputenc}
\usepackage{booktabs}
\usepackage{tablefootnote}

\usepackage{microtype}

\usepackage{inconsolata}

\usepackage{graphicx}

\newcommand*{\affmark}[1][*]{\textsuperscript{#1}}
\title{WisPerMed at BioLaySumm: Adapting Autoregressive Large Language Models for Lay Summarization of Scientific Articles}

\author{\bf Tabea M. G. Pakull\affmark[1,2],\quad Hendrik Damm\affmark[2,3], \quad \bf Ahmad Idrissi-Yaghir\affmark[2,3],\\ \bf \quad Henning Schäfer\affmark[1,2],\quad \bf Peter A. Horn\affmark[1], and Christoph M. Friedrich\affmark[2,3]\\
\affmark[1] Institute for Transfusion Medicine, University Hospital Essen,\\ Hufelandstraße 55, 45147 Essen, Germany\\
\affmark[2] Department of Computer Science, University of Applied Sciences and Arts Dortmund, \\ Emil-Figge-Straße 42, 44227 Dortmund, Germany\\
\affmark[3] Institute for Medical Informatics, Biometry and Epidemiology (IMIBE), \\ University Hospital Essen, Hufelandstraße 55, 45147 Essen, Germany\\
\texttt {tabeamargaretagrace.pakull@uk-essen.de}}

\begin{document}
\maketitle
\begin{abstract}
This paper details the efforts of the WisPerMed team in the BioLaySumm2024 Shared Task on automatic lay summarization in the biomedical domain, aimed at making scientific publications accessible to non-specialists. Large language models (LLMs), specifically the BioMistral and Llama3 models, were fine-tuned and employed to create lay summaries from complex scientific texts. The summarization performance was enhanced through various approaches, including instruction tuning, few-shot learning, and prompt variations tailored to incorporate specific context information. The experiments demonstrated that fine-tuning generally led to the best performance across most evaluated metrics. Few-shot learning notably improved the models' ability to generate relevant and factually accurate texts, particularly when using a well-crafted prompt. Additionally, a Dynamic Expert Selection (DES) mechanism to optimize the selection of text outputs based on readability and factuality metrics was developed. 
Out of 54 participants, the WisPerMed team reached the 4th place, measured by readability, factuality, and relevance. Determined by the overall score, our approach improved upon the baseline by $\approx5.5$ percentage points and was only $\approx1.5$ percentage points behind the first place.
\end{abstract}
\section{Introduction}
In the biomedical domain, scientific publications and research play a central role in communicating research findings and results. However, these documents are usually written in complex language and use terminology and technical jargon that can be challenging for lay readers or researchers from different fields to understand \citep{goldsack_making_2022}. In this context, lay summarization can be utilized to extract the most relevant information from the original article or publication while also providing supplementary explanations. This often entails incorporating background information that may not be contained within the article itself.

In this context, this paper presents the participation of the team WisPerMed in the BioLaySumm2024 Shared Task \citep{goldsack_overview_2024} on automatic lay summarization and describes the employed approaches to tackle this challenge.

Summaries generated by LLMs, as demonstrated by \citet{zhang_benchmarking_2024}, can be of equivalent or superior quality to original references. Additionally, instruction tuning is an effective approach for enhancing performance. However, LLMs face limitations when applied to domain-specific abstractive summarization. Key challenges include the quadratic complexity of transformer-based models \citep{ashish_vaswani_attention_2017} concerning input text length, model hallucination, where factually incorrect text is generated, and domain shift from training to test data \citep{afzal_challenges_2023}. Similarly, studies on text simplification \citep{amin_artificial_2023} indicate that although general-purpose LLMs are capable of effectively simplifying clinical reports, they sometimes generate factual inaccuracies and omit crucial information. 

To adapt LLMs to a specific domain or task \citep{ling_domain_2024}, it is possible to fine-tune the models, leverage few-shot learning or further pre-train the models on domain data. Examples of domain-adapted LLMs for the biomedical domain include the BioMistral \citep{labrak_biomistral_2024} and OpenBioLLM \citep{pal_openbiollms_2024} model series. The BioMistral models are based on the Mistral 7B Instruct v0.1 \citep{jiang_mistral_2023} model. They are further pre-trained on the PMC Open Access Subset\footnote{\url{https://www.ncbi.nlm.nih.gov/pmc/tools/openftlist/} Accessed: 2024-05-17}. OpenBioLLM models are based on the Llama3  \citep{aimeta_llama_2024} models and were adapted to the biomedical domain through fine-tuning.

\section{Dataset}
The dataset \citep{goldsack_making_2022} of the Shared Task \citep{goldsack_overview_2024} contains two collections of scientific journal articles and the corresponding lay summaries, namely PLOS and eLife. PLOS and eLife also include the section headings and keywords of the article. The PLOS dataset has 24,773 examples in the training split and 1,376 examples in the validation split, whereas the eLife dataset is smaller with 4,346 examples in the training split and 241 examples in the validation split. The test split consists of 142 examples for both datasets. Lay summaries of the PLOS dataset were written by the authors of the articles and are approximately 150-200 words long, while eLife lay summaries were written by expert editors in correspondence with the authors and are about twice as long.

For the remainder of this paper, any reference to the validation or test set will include eLife and PLOS unless otherwise specified.

\section{Evaluation Metrics}
The generated summaries were evaluated across ten metrics that fall into the following categories: relevance, readability, and factuality. Relevance was assessed through Recall-Oriented Understudy for Gisting Evaluation \citep{lin_rouge_2004} (ROUGE-1, ROUGE-2, ROUGE-L) and BERTScore \citep{zhang_bertscore_2020}. ROUGE counts the overlapping n-grams in the generated texts and target lay summaries, whereas BERTScore uses contextual word embeddings to compare the semantic similarity of the two texts. Readability was evaluated using the Flesch-Kincaid Grade Level (FKGL) \citep{kincaid_derivation_1975}, Dale-Chall Readability Score (DCRS) \citep{chall_readability_1995}, Coleman-Liau Index (CLI) \citep{coleman_computer_1975}, and Learnable Evaluation Metric for Text Simplification (LENS) \citep{maddela_lens_2023}. The FKGL uses sentence lengths and syllable count per word to estimate readability. The DCRS uses a word list to compute the occurrences of words unknown to most 4-th graders and the CLI estimates the grade level necessary to comprehend the text. The LENS metric is a learnable evaluation metric trained on datasets containing human ratings of simplifications. In this setting, LENS measures the simplification of the abstract by the generated text using the target lay summary as a reference. Factuality was assessed with AlignScore \citep{zha_alignscore_2023} and Summary Consistency (SummaC) \citep{laban_summac_2022}. The AlignScore quantifies the degree of alignment between the facts in the summary and the scientific article, while SummaC also includes consistency.
\section{Methods and Experiments}
This section outlines the methodology employed in the experiments conducted on the specified dataset.
\begin{figure*}[htbp]
  \centering
\includegraphics[scale=0.9]{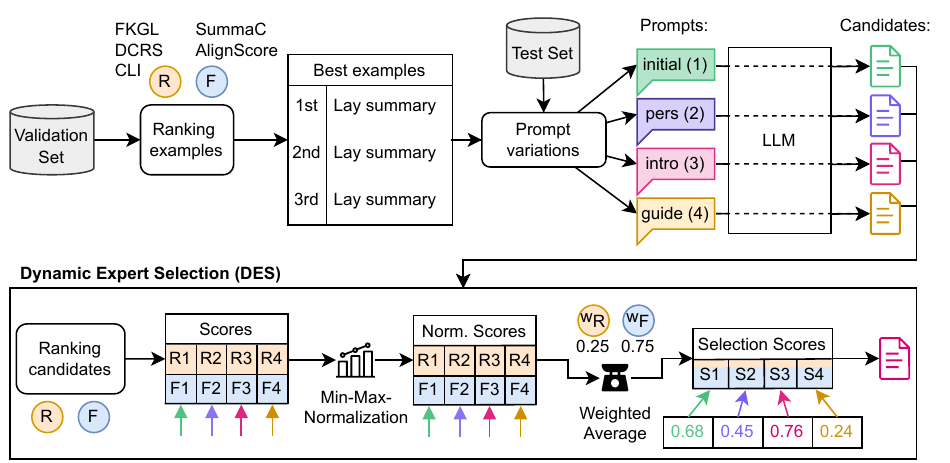}  \caption{Workflow of the Dynamic Expert Selection (DES) mechanism in the few-shot setting using an example from the PLOS dataset. The process involves ranking examples, generating multiple summaries through various prompt variations, applying a large language model (LLM), and then normalizing and weighing the readability (R) and factuality (F) scores to rank and select the best summary based on the selection scores (S).}
\label{fig:work}
\end{figure*}

\subsection{Fine-tuned Models}
In this study, instruction tuning \citep{wei_finetuned_2022} was utilized to fine-tune various models. Instruction tuning refers to the process of fine-tuning language models on a collection of datasets described via instructions. BioMistral-7B-DARE (BioM) and Llama3-70B-Instruct (Llama3) were fine-tuned for one epoch utilizing Quantized Low-Rank Adaptation (QLoRA) \citep{dettmers_qlora_2023} on the eLife and PLOS dataset individually. BioM was trained on the abstracts + lay summaries, whereas Llama3 was trained on the entire articles + lay summaries. The texts were structured using the Mistral and Llama3 instruction templates prior to the fine-tuning process. Please refer to the Appendix \ref{app:prompt}, \ref{app:hyper}, and \ref{sec:appendix_lic} for details on prompts, parameters and licenses, respectively.

After evaluating the checkpoints of BioM on the validation set, the checkpoints with the best scores were selected for inference. For Llama3, the final checkpoints were selected. The models were given the same prompt as during fine-tuning but without the target. 
\subsection{Prompt Variations}
Prompts can guide the LLM’s content generation process without the need for fine-tuning. In the zero- and few-shot settings, different prompt variations and their effect on the evaluation metrics were examined. In the few-shot setting, example lay summaries from the training and validation set were included in the prompt when performing inference on the validation and test set, respectively. The format of these few-shot prompts is designed to emulate a preceding conversation with the model, with the included examples serving as the model's previous responses.

To choose the best few-shot examples, all examples were ranked based on their average normalized readability and factuality. The two and three highest-ranked examples were selected for the eLife and PLOS datasets, respectively.

An initial prompt was created by replicating the prompt used for inference with the fine-tuned BioM model (see Appendix \ref{app:prompt}). This prompt was then tested with BioM and OpenBioLLM-70B (OpenBio) on the validation set.

Additionally, three prompt variations were created, which provide the model with different kinds of context information. It was decided that BioM would be utilized for all experiments involving these variations due to its superior performance on the validation set in the few-shot setting (see Table \ref{valid-results} in Appendix \ref{app:valid}). LLMs can assume different roles and adapt their vocabulary accordingly \citep{salewski_-context_2023}, resulting in enhanced performance in tasks related to the specified role. Accordingly, the first prompt variation comprises a persona description of a science communicator (BioM\textsubscript{pers}), instructing the model to utilize the expertise of this persona to create the lay summary based on the abstract. The model is then instructed to channel the expertise of the described persona to craft the lay summary based on the abstract. The second prompt variation is a modification of the initial prompt, incorporating the introduction to provide additional background information because associated context can improve LLM performance \citep{karmaker_santu_teler_2023}. The second prompt variation is a modification of the initial prompt, but it includes the introduction as further context for background information (BioM\textsubscript{intro}). The third prompt variation includes the abstract and a guide on how to write a lay summary (BioM\textsubscript{guide}), accompanied by instructions concerning the content and style of the requested summary. This method leverages the importance of clear and detailed task directives. The selection of these prompts was based on a few preliminary experiments with the model and an initial assessment of the responses. However, no comprehensive optimization was performed. The wording of all prompts can be found in Appendix \ref{app:prompt}.

Due to the efficacy of few-shot learning with the initial prompt, the prompt variations were implemented in a few-shot setting on the test set.

\subsection{Dynamic Expert Selection (DES)}
The success of an LLM depends on factors such as the properties of the dataset, the complexity of the domain, and the design of the prompt \citep{ling_domain_2024}. Consequently, a model may yield a more suitable lay summary when prompted in a different manner. In addition, the output quality depends upon the selection of the inference parameters \citep{minaee_large_2024}. In consideration of this assumption, a Dynamic Expert Selection (DES) was developed. It selects the most appropriate text from a set of candidate texts based on metrics that do not require a reference lay summary. 

The mechanism uses the readability metrics FKGL, DCRS, and CLI, as well as the factuality metrics AlignScore and SummaC. These metrics are computed for each candidate text. Readability scores are multiplied by -1 so that higher scores indicate better readability. All scores are normalized using min-max normalization to range between 0 and 1, where 1 is the best and 0 is the worst. For each candidate text, an overall score is calculated by multiplying the means with different weights. Given that the target lay summaries in eLife have a higher readability than those in PLOS \citep{goldsack_making_2022}, the overall scores are computed with different weights for the two aspects. For eLife summaries: Readability is weighted at 0.675 and factuality at 0.325. For PLOS summaries: Readability is weighted at 0.25 and factuality at 0.75. The candidate text with the highest overall score is selected as the most suitable lay summary. The selection of the weights is based on the assumptions about the target texts and comparisons of the overall scores on the validation dataset.

This approach was applied to BioM in the few-shot setting using all prompt variants (see Figure \ref{fig:work}) and to the fine-tuned BioM using two distinct inference parameter settings (see Appendix \ref{app:hyper}).

\begin{table*}
\centering

\begin{tabular}{ l  l  l  l  l  l  l  l  l  l  l } 
\toprule    
\textbf{Expt.} & \textbf{R-1}& \textbf{R-2}& \textbf{R-L}& \textbf{BERT}& \textbf{FKGL} & \textbf{DCRS} & \textbf{CLI} & \textbf{LENS} & \textbf{Align}& \textbf{SC}\\
\midrule
Baseline& 0.470&0.140& 0.436& 0.862& 12.036& 10.148& 13.485& 48.096& \underline{0.779}&0.703\\
  \midrule
 \multicolumn{11}{c}{Zero-shot Learning}\\
 \midrule
 BioM& 0.329 & 0.071 & 0.298 & 0.845 & 12.404 & 10.093 & 13.974 & \underline{80.396}& 0.541 &0.458 \\
 \midrule
 \multicolumn{11}{c}{Few-shot Learning}\\
 \midrule
BioM& 0.440 & 0.124 & 0.409 & \textbf{0.857} & 11.287 &\textbf{8.954} & \underline{\textbf{12.755}}& 75.744 & 0.728 &0.604 \\
BioM\textsubscript{pers}& \textbf{0.442} & 0.125 & \textbf{0.412} & 0.856 & 11.318 & 9.066 & 13.031 & 63.766 & 0.721 &0.607 \\
BioM\textsubscript{intro}& 0.391 & 0.106 & 0.359 & 0.851 & 12.233 & 9.618 & 13.693 & 76.638 & 0.669 &0.529 \\
BioM\textsubscript{guide}& 0.434 & 0.117 & 0.403 & 0.856 & 11.773 & 9.553 & 13.662 & 76.912 & 0.692 &0.557 \\
BioM\textsubscript{DES}& 0.439 & \textbf{0.128} & 0.409 & 0.855 & \underline{\textbf{10.969}} & 8.993 & 12.819 & 74.025 & \textbf{0.767} & \textbf{0.673} \\
OpenBio& 0.415 & 0.104 & 0.382 & 0.855 & 11.657 & 9.848 & 13.711 & \textbf{79.519} & 0.731 &0.558 \\

\midrule
 \multicolumn{11}{c}{Fine-tuning}\\
 \midrule
BioM& 0.470 & \underline{\textbf{0.152}} & 0.442 & \underline{\textbf{0.865}} & 11.338 & 8.872 & 13.064 & 51.058 & 0.775 & 0.705 \\
BioM\textsubscript{DES}& \underline{\textbf{0.471}} & \underline{\textbf{0.152}} & \underline{\textbf{0.443}} & \underline{\textbf{0.865}} & \textbf{11.072} & \underline{\textbf{8.862}} & \textbf{12.871} & 51.028 & \textbf{0.782} & \underline{\textbf{0.722}} \\
Llama3& 0.418 & 0.108 & 0.391 & 0.856 & 11.622 & 10.628 & 15.080 & \textbf{72.860} & 0.602 & 0.592 \\

\bottomrule

\end{tabular}
\caption{\label{test-results}Performance metrics of experiments on the test set. The models include BioMistral-7B (BioM), Llama3-70B (Llama3), and OpenBioLLM-70B (OpenBio). The experiments are categorized into fine-tuned, zero-shot, and few-shot settings. The metrics reported are ROUGE-1 (R-1), ROUGE-2 (R-2), ROUGE-L (R-L), BERTScore (BERT), FKGL, DCRS, CLI, LENS, AlignScore (Align), and SummaC (SC). Bolded values indicate the best in each section, and underlined values the best overall performance.}

\end{table*}
\section{Results}
The results of the experiments using BioM, Llama3, and OpenBio are presented in table \ref{test-results}. The experiments are categorized into zero-shot learning, few-shot learning, and fine-tuning.

BioM exhibits the highest LENS score in the zero-shot setting. However, its relevance and factuality performance are the lowest. Few-shot learning resulted in enhanced performance across all metrics except for LENS. The persona prompt (BioM\textsubscript{pers}) led to an improvement in relevance. Including the introduction in the prompt (BioM\textsubscript{intro}) resulted in a reduction in all aspects despite the fact that the model had access to more information from the article itself. In comparison, the prompt with the guide (BioM\textsubscript{guide}) exhibits minimal enhancements. The optimal few-shot learning for BioM occurred with the initial prompt, which achieved the highest readability and factuality in the few-shot setting, excluding the DES approach. However, OpenBio slightly underperformed with this prompt in the few-shot setting, except for the LENS score, where it performed best in this setting.

The DES used all four prompts and outperformed the baseline with improvements in factuality and readability, achieving the best results in the few-shot setting.

Fine-tuning BioM improved relevance and factuality scores, though the LENS score decreased slightly, with other readability metrics similar to the few-shot setting. The fine-tuned BioM outperformed the baseline in terms of relevance and overall quality. The DES approach improved all metrics except for a slight drop in the LENS score. In contrast, Llama3 underperformed despite being larger. It was less effective at extracting relevant information from full articles and produced lower-quality text in terms of readability, even though its LENS score was higher than BioM's. Additionally, Llama3's factuality scores decreased, leading to an overall performance drop compared to the baseline.

\section{Conclusion}
This paper presents the WisPerMed team's approaches to automatic lay summarization within the biomedical domain, utilizing a combination of fine-tuning, prompt variations, and Dynamic Expert Selection.

Among these approaches, fine-tuning emerged as an effective method, leading to the best performance across most metrics. This underscores the importance of task-specific training in optimizing model output for complex summarization tasks. Additionally, BioM showed strong few-shot learning capabilities, illustrating its robustness and versatility in generating accurate and relevant summaries even without extensive training. As the model adjusts to the factuality and readability of given examples, providing better examples could lead to further enhancements in these aspects. 

BioM reached high factuality, even when provided solely with abstracts as input, suggesting that BioM leveraged domain-specific knowledge acquired during pre-training. This indicates that domain adaptation remains an important factor when using LLMs for lay summarization of scientific articles, as BioM outperformed the larger general model Llama3.  

The four prompt variations exhibited differing effects on the evaluation metrics. BioM is adept in fulfilling the role of a science communicator (BioM\textsubscript{pers}), as evidenced by the enhanced relevance. BioM\textsubscript{intro} and BioM\textsubscript{guide} did not significantly enhance the metrics, indicating that the increase in context was not beneficial for all texts. Without DES, a shorter prompt (BioM\textsubscript{initial}) yielded the optimal results, suggesting that the model effectively comprehends the task from the provided examples. The DES mechanism further refined readability and, in particular, factuality by retrospectively selecting the best text outputs based on evaluation metrics. This highlights the potential of metric-driven selection to improve the quality of lay summaries further.

In conclusion, our study demonstrates that fine-tuning, the use of informed prompt variations, and selection mechanisms can enhance the capability of autoregressive LLMs to produce lay summaries that are factually accurate, relevant, and readily accessible to non-specialist audiences. This approach fosters broader public engagement with scientific findings, advancing the goal of making biomedical research comprehensible and accessible.
\section*{Limitations}
Only four discrete prompts in combination were tested with DES, and only two sets of inference parameters were explored. This limited scope means that the findings may not fully capture the potential variability and performance of the the various models under different conditions.
The weights for the Dynamic Expert Selection method were chosen based on heuristics without any formal optimization, which could impact the robustness and generalizability of the results. Another limitation is the possibility that BioM may have been previously exposed to the gold standard summaries. If this is the case, it could skew the results by artificially inflating the model's performance. These limitations indicate potential avenues for future research, including the necessity for more comprehensive prompt engineering, optimization of DES weights, and a wider range of tasks to ensure the robustness of the approach. Another potential future direction is adapting these methods for other complex domains or languages and exploring additional metrics.
\section*{Acknowledgement}
The work of Tabea M. G. Pakull, Hendrik Damm, Ahmad Idrissi-Yaghir and Henning Schäfer was funded by a PhD grant from the DFG Research Training Group 2535 \textit{Knowledge-and data-based personalisation of medicine at the point of care (WisPerMed)}.

\bibliography{acl_latex}

\onecolumn
\appendix
\section{Prompts}
\label{app:prompt}
The prompts used in the experiments are shown in Figures \ref{fig:ini}, \ref{fig:llama3}, \ref{fig:persona}, \ref{fig:intro}, and \ref{fig:guide}.
\begin{figure}[hbtp]
     \begin{tcolorbox}[boxrule=0.5pt, colframe=black,coltitle=white, colbacktitle=darkgray, width=\textwidth, title=Fine-tuning/Initial Prompt with Abstract for BioM]
    {
    \footnotesize        
        
        You will be provided with the abstract of a scientific article. Your task is to write a lay summary that accurately conveys the key findings and significance of the research in non-technical language understandable to a general audience.:
        
        \vspace{0.4em}

        Abstract of a scientific article: 
        \vspace{0.4em}

        [Abstract]
        
        \vspace{0.4em}

        Lay summary for this article:

        }
        
    \end{tcolorbox}
    \caption{The prompt used for fine-tuning BioM and as the initial prompt in the zero- and few-shot settings. For fine-tuning the prompt also includes the target lay summary.}
        \label{fig:ini}

\end{figure}

\begin{figure}[hbtp]
     \begin{tcolorbox}[boxrule=0.5pt, colframe=black,coltitle=white, colbacktitle=darkgray, width=\textwidth, title=Fine-tuning/Inference Prompt with Article for LLama3]
    {
    \footnotesize        
        
        You will be provided with a scientific article. Your task is to write a lay summary that accurately conveys the key findings and significance of the research in non-technical language understandable to a general audience.:
        
        \vspace{0.4em}

        Scientific article: 
        \vspace{0.4em}

        [Abstract]
        
        \vspace{0.4em}

        Lay summary for this article:

        }
        
    \end{tcolorbox}
    \caption{The prompt used for fine-tuning Llama3. For fine-tuning the prompt also includes the target lay summary.}
        \label{fig:llama3}

\end{figure}
\begin{figure}[hbtp]
     \begin{tcolorbox}[boxrule=0.5pt, colframe=black,coltitle=white, colbacktitle=darkgray, width=\textwidth, title=Persona Prompt]
    {
    \footnotesize        
        
         Meet Layla, your fantastic science communicator committed to breaking down complex research for everyone! Layla's mission is to create summaries that make scientific literature easy to understand for the general public. Before writing, Layla thoroughly reads the abstract to grasp the research goals and findings accurately. Precision is crucial for Layla; she makes sure her summaries align with the abstract's research while expanding on key points and methods. Layla ensures each summary gives a complete understanding of the findings and their importance. She offers detailed explanations and backgound information as context to aid comprehension. She highlights the main discoveries and their real-world implications, explaining study mechanisms and methods in reader-friendly language. Layla brings research to life with vivid descriptions and relatable examples, showing its impact on society. Her tone is informative yet engaging, avoiding jargon to be inclusive.
        \vspace{0.4em}
      
                Now, let's channel Layla's expertise to craft a comprehensive lay summary for a scientific article.

        \vspace{0.4em}

                Abstract of the scientific article:
                        \vspace{0.4em}

        [Abstract]
                \vspace{0.4em}

        Layla:
            
        }
        
    \end{tcolorbox}
    \caption{The Persona-Prompt used in zero- and few-shot setting with BioM.}
        \label{fig:persona}

\end{figure}
\begin{figure}[hbtp]
     \begin{tcolorbox}[boxrule=0.5pt, colframe=black,coltitle=white, colbacktitle=darkgray, width=\textwidth, title=Intro Prompt]
    {
    \footnotesize

        You will be provided with the abstract of a scientific article and the introduction section for background information. Your task is to write a lay summary that accurately conveys the key findings and significance of the research in non-technical language understandable to a general audience. Please ensure that your summary is mainly based on the information provided in the abstract. You may also use information from the introduction for additional context if necessary.

        \vspace{0.4em}

        Introduction of the scientific article:
        \vspace{0.4em}

        [Introduction]

        \vspace{0.4em}
        
        Abstract of the scientific article:

        \vspace{0.4em}

        [Abstract]

                \vspace{0.4em}

        Lay summary for this article:\\
        
        }
        
    \end{tcolorbox}
    \caption{The Intro-Prompt used in zero- and few-shot setting with BioM.}
        \label{fig:intro}

\end{figure}

\begin{figure}[hbtp]
     \begin{tcolorbox}[boxrule=0.5pt, colframe=black,coltitle=white, colbacktitle=darkgray, width=\textwidth, title=Guide Prompt]
    \footnotesize        
        
You will be provided with the abstract of a scientific article. Your task is to write a lay summary that accurately conveys the key findings and significance of the research in non-technical language understandable to a general audience.
        \vspace{0.4em}
  
Abstract of the scientific article: 
        \vspace{0.4em}

[Abstract]
        \vspace{0.4em}

Guidelines for crafting a lay summary:
\begin{itemize}
    \item Craft a detailed summary that explains the research findings and their implications, providing thorough explanations where necessary.
    \item Ensure factual accuracy and alignment with the research presented in the abstract, elaborating on key points and methodologies.
    \item Highlight the main findings and their implications for real-world scenarios, delving into specific mechanisms or methodologies used in the study and their broader significance.
    \item Incorporate descriptive language to explain complex concepts.
    \item Maintain a balanced tone that is informative and engaging, avoiding technical jargon or overly formal language.
    \item Ensure the summary provides sufficient depth and context to guide the reader through the research journey and address potential questions or areas of confusion.
\end{itemize}
        \vspace{0.4em}

Your lay summary for the article:\\
            
    \end{tcolorbox}
    \caption{The Guide-Prompt used in zero- and few-shot setting with BioM.}
    \label{fig:guide}
\end{figure}
\newpage
\section{Setup and Hyperparameter}
\label{app:hyper}

\paragraph{Training}All trainings were executed on a single Nvidia H100 80GB using the unsloth\footnote{\url{https://github.com/unslothai/unsloth} Accessed: 2024-05-17} framework and QLoRA \citep{dettmers_qlora_2023}. The following modules were targeted with QLoRA: ``q\_proj'', ``k\_proj'', ``v\_proj'', ``o\_proj'', ``gate\_proj'', ``up\_proj'', and ``down\_proj''. The QLoRA rank and alpha were both set to 16. The QLoRA dropout was set to 0. The optimization of the models was conducted using the 8-bit Adam optimizer \citep{loshchilov_decoupled_2019}, which was configured with a maximum learning rate of $2 \times 10^{-4}$ and a weight decay factor of 0.01. The learning rate schedule included a linear decay following an initial phase consisting of five warm-up steps. Maximum sequence length was set to 4,096.

\paragraph{Inference}For the inference process, a greedy search algorithm was employed as the decoding strategy \citep{minaee_large_2024}, with a configuration that allowed for the generation of up to 1024 new tokens per inference iteration.

\paragraph{DES}The DES with the fine-tuned model used the inference parameter as described above for one candidate, and a repetition penalty of 1.1 was chosen to generate another candidate.

\section{Licenses}
\label{sec:appendix_lic}
In Table \ref{tab:translation_metrics} the Licenses as given by the owners of the Framework/Model are displayed.

\begin{table}[htbp]
\centering
\begin{tabular}{llll}
\toprule
Framework/Model                   & License\\ 
\midrule
unsloth\tablefootnote{\url{https://github.com/unslothai/unsloth} Accessed: 2024-05-17}                  & Apache License Version 2.0\\

BioMistral-7B-DARE\tablefootnote{\url{https://huggingface.co/BioMistral/BioMistral-7B-DARE} Accessed: 2024-05-17}                & Apache License Version 2.0\\
Llama-3-70B-I\tablefootnote{\url{https://huggingface.co/meta-llama/Meta-Llama-3-70B-Instruct} Accessed: 2024-05-17}                & Llama 3 Community License Agreement\\
OpenBioLLM-70B\tablefootnote{\url{https://huggingface.co/aaditya/Llama3-OpenBioLLM-70B} Accessed: 2024-05-17}                & Llama 3 Community License Agreement\\
\bottomrule
\end{tabular}
\caption{Licenses of the dataset, Framework and Models used for this Shared Task.}
\label{tab:translation_metrics}
\end{table}
\newpage
\section{Results on the Validation Set}
\label{app:valid}
The results of experiments on the validation set and the reference scores of the target lay summaries and input abstracts are presented in Table \ref{valid-results}.

\begin{table*}[htbp]
\centering
\begin{tabular}{ l  l  l  l  l  l  l  l  l  l  l }  
\toprule
\textbf{Expt.}& \textbf{R-1}& \textbf{R-2}& \textbf{R-L}& \textbf{BERT}& \textbf{FKGL} & \textbf{DCRS} & \textbf{CLI} & \textbf{LENS} & \textbf{Align}& \textbf{SC}\\
\midrule
Targets& -& -& -& -& 12.857 & 9.944 & 14.251 & 57.988 & 0.670 & 0.512 \\
 Abstracts& 0.410 & 0.135 & 0.380 & 0.855 & 15.260 & 11.378 & 16.961 & 38.259 & -&-\\
\midrule
 \multicolumn{11}{c}{Zero-shot Learning}\\

 \midrule
BioM& 0.332 & 0.070 & 0.301 & 0.844 & \textbf{12.530} & 10.156 &\textbf{13.957} & \underline{\textbf{80.159}} & 0.521 & 0.465 \\
BioM\textsubscript{pers}& 0.411 & 0.118 & 0.379 & 0.847 & 12.579 & \textbf{10.074} & 14.897 & 69.732 & 0.741 & 0.628 \\
BioM\textsubscript{intro}& 0.397 & 0.118 & 0.364 & 0.849 & 13.735 & 10.478 & 14.990 & 68.530 & 0.743 & 0.580 \\
BioM\textsubscript{guide}& \textbf{0.422} &\textbf{0.123} & \textbf{0.389} & \textbf{0.851} & 13.971 & 10.478 & 15.667 & 68.561 & \underline{\textbf{0.747}} & \textbf{0.593} \\

 \midrule

 \multicolumn{11}{c}{Few-shot Learning}\\
\midrule
BioM& \textbf{0.440} & \textbf{0.122} & \textbf{0.411} & \textbf{0.855} & \underline{\textbf{10.875}}& \underline{\textbf{8.733}}& \underline{\textbf{12.359}}& 76.358 & \textbf{0.701} & \textbf{0.596} \\
OpenBio& 0.423 & 0.107 & 0.390 & 0.854 & 12.429 & 9.729 & 14.721 & \textbf{77.961} & 0.678 & 0.554 \\
 \midrule
 \multicolumn{11}{c}{Fine-tuning}\\
\midrule
BioM& \underline{0.478} & \underline{0.148} & \underline{0.446} & \underline{0.866} & 11.743 & 9.899 & 13.886 & 56.888 & 0.724 & \underline{0.677}\\
\bottomrule

\end{tabular}

\caption{\label{valid-results} Performance metrics of experiments on the validation set. The models include BioMistral-7B (BioM), Llama3-70B (Llama3), and Llama3-OpenBioLLM-70B (OpenBio). The experiments are categorized into fine-tuned, zero-shot, and few-shot settings. The metrics reported are ROUGE-1 (R-1), ROUGE-2 (R-2), ROUGE-L (R-L), BERTScore (BERT), FKGL, DCRS, CLI, LENS, AlignScore (Align), and SummaC (SC). 'Targets' and 'Abstracts' provides benchmark scores of the target lay summaries and abstracts, respectively. Bolded values indicate the best in each section, and underlined values the best overall performance.}
\end{table*}

\end{document}